\documentclass[letterpaper]{article} 
\usepackage{aaai2027} 
\usepackage[hyphens]{url}  
\usepackage{graphicx} 
\usepackage{natbib}  
\usepackage{caption} 
\usepackage{algorithm}
\usepackage{algorithmic}
\usepackage{multirow}
\usepackage{amsmath}
\usepackage{xcolor}
\usepackage{colortbl}
\usepackage{makecell}
\usepackage{newfloat}
\usepackage{listings}
\DeclareCaptionStyle{ruled}{labelfont=normalfont,labelsep=colon,strut=off} 
\floatstyle{ruled}
\newfloat{listing}{tb}{lst}{}
\floatname{listing}{Listing}

\usepackage{booktabs}

\title{Rewarding Reasoning, Not Answers: Fixing and Bounding\\ Test-Time Reinforcement Learning on Medical QA}
\author{
    Kailong Fan\textsuperscript{\rm 1,\rm 2},
    Anqi Pu\textsuperscript{\rm 3},
    Yichen Wu\textsuperscript{\rm 1}\corresponding,
    Wanhua Li\textsuperscript{\rm 3},
    Yicong Li\textsuperscript{\rm 3},\\
    Hanspeter Pfister\textsuperscript{\rm 3},
    Huafeng Liu\textsuperscript{\rm 2},
    Xiang Li\textsuperscript{\rm 1},
    Quanzheng Li\textsuperscript{\rm 1},
    Ning Guo\textsuperscript{\rm 1}
}

\affiliations{
    \textsuperscript{\rm 1}Harvard Medical School/MGH\\
    \textsuperscript{\rm 2}Zhejiang University\\
    \textsuperscript{\rm 3}Harvard University\\
    Corresponding author: yiwu6@mgh.harvard.edu
}

\begin{document}

\nocopyright  
\maketitle

\begin{abstract}
Test-time reinforcement learning adapts a model on its own unlabeled test
set using majority-vote pseudo-labels and has shown strong results in mathematics. We show that this recipe collapses on medical multiple-choice QA: accuracy stagnates while output diversity rapidly declines. Through a controlled experiment that keeps the questions, model, and optimizer fixed while changing only the answer space, we trace this failure to answer-space structure rather than domain difficulty. In small answer spaces, incorrect rollouts often collide on the same wrong pseudo-label and reinforce it; in large answer spaces, they disperse and receive little reward. This diagnosis motivates PROSE, Process Reward Guided Self-Training, which rewards reasoning quality instead of answer agreement. PROSE scores each reasoning step with a medical process reward model, assigns the trajectory reward as the minimum score across steps, and enforces answer-format constraints. Without labels, PROSE substantially improves a general Llama model, surpassing purpose-built medical models and matching much larger systems. Because the process signal is internalized into the policy, the adapted model requires no reward model at inference and transfers its gains to unseen datasets. We further show that the minimum aggregation is essential: mean aggregation can be exploited, saturating the proxy reward while degrading accuracy.

\end{abstract}

\section{Introduction}
\label{sec:intro}

\begin{figure}[t]
  \centering
  \includegraphics[width=\columnwidth]{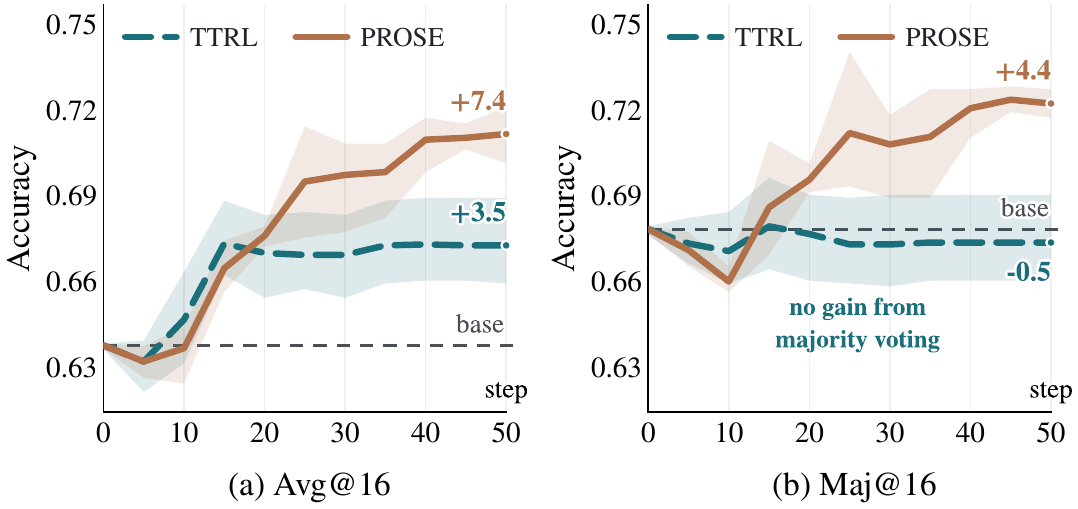}
\caption{\textbf{Rewarding reasoning prevents test-time RL collapse.}
Standard TTRL yields only modest gains in average accuracy over 16 sampled
responses (avg@16) \textbf{(a)} and no gain in majority-vote accuracy over
16 sampled responses (maj@16) \textbf{(b)}, whereas PROSE consistently
improves both metrics. Curves show the mean, and shaded
regions denote $\pm$ one standard deviation across three random seeds.}
  \label{fig:teaser}
\end{figure}

Recent work shows that large language models can adapt at test time using their own sampled outputs~\cite{zuo2026ttrl,Moradi2025ContinuousSO,yang2026ttcs}. Test-time reinforcement learning (TTRL) turns these outputs into a learning signal: for each unlabeled test question, the model samples multiple responses, treats the majority answer as a pseudo-label, and updates itself to agree with it. This simple recipe has produced strong gains in mathematics, making it attractive since it requires only test questions rather than ground-truth labels or additional annotation.

Medical question answering is a natural setting where such label-free adaptation would be especially valuable. Expert annotation is costly, clinical distributions vary across institutions and specialties, and privacy constraints often limit external labeling~\cite{wang2021annotation,zech2018variable,price2019privacy}. At the same time, many medical benchmarks and decision-support interfaces are formulated as multiple-choice QA (MCQ), where the answer space is small and highly structured. This format raises a central question: does majority-vote self-training remain reliable when many different reasoning paths must choose from the same limited set of answers?




Figure~\ref{fig:teaser} illustrates the empirical failure that motivates this work. Standard TTRL yields only marginal improvement in sample-level average accuracy and no improvement in majority-vote accuracy, despite repeated adaptation on the unlabeled test set. At the same time, the model's outputs become substantially less diverse, indicating that adaptation increases agreement among samples without improving correctness. In contrast, our PROSE improves both accuracy metrics by replacing final-answer agreement with process-level reward. This contrast suggests the limitation lies not in test-time adaptation itself, but in using majority-vote pseudo-labels within a small, structured answer space.

To isolate the source of this collapse, we conduct a controlled experiment that holds the questions, model, optimizer, and training procedure fixed, and varies only the answer space. The results support an answer-space explanation rather than a domain-difficulty explanation. We show that the \emph{Lucky Hit} mechanism identified by TTRL~\cite{zuo2026ttrl} depends on the structure of the answer space: in a large free-form answer space, incorrect rollouts usually disperse across distinct strings and are rarely reinforced by the majority-vote reward; in a small multiple-choice answer space, different mistaken rationales can converge on the same wrong option, producing a false consensus that is then rewarded as a pseudo-label. We quantify this effect through the protection gap between pseudo-label reward accuracy and true answer accuracy, and find that the gap present under free-form answering nearly vanishes under multiple choice.

This diagnosis points to a different training signal. If agreement on the final answer is unreliable, the reward should evaluate the reasoning process that produced the answer~\cite{uesato2022solving,Lightman2023Let's}. We introduce PROSE, which keeps the test-time RL loop but replaces majority-vote pseudo-labels with a step-level medical process reward model. PROSE scores each reasoning step, sets the trajectory reward to the minimum step score, and applies an answer-format guard. Minimum aggregation makes the reward depend on the weakest step in the chain, preventing trajectories from hiding flawed reasoning behind many easy or fluent steps.

Beyond improving adaptation, PROSE has a practical deployment advantage. The PRM is used only during the adaptation stage to provide process-level rewards for sampled trajectories. After adaptation, the policy is evaluated on its own, using ordinary sampling or majority voting without access to the PRM. Thus, the process signal is absorbed into the model parameters rather than applied as an external selector at inference time. Empirically, this learned policy outperforms PRM-based selection under the same sampling budget and preserves its gains on datasets that were never used for adaptation, indicating that PROSE learns transferable reasoning behavior rather than dataset-specific answer preferences. To sum up, our contributions are:
\begin{itemize}
    \item We provide a causal account of why TTRL fails on medical multiple-choice QA. The Lucky Hit mechanism shows how small answer spaces convert independent reasoning errors into reinforced pseudo-label errors.
    \item We introduce PROSE, a label-free test-time adaptation method that rewards reasoning quality with a medical process reward model, assigns trajectory rewards by the minimum score across reasoning steps, and guards the answer format.
    \item We show that PROSE internalizes the process reward into the policy, requires no reward model at inference, generalizes beyond the adapted datasets, and avoids the reward hacking observed with mean aggregation.
\end{itemize}

\section{Related Work}
\label{sec:related} 
\noindent \textbf{TTS and Self-Evolution.} Increasing computation at inference time has become a practical way to elicit stronger reasoning from LLMs, often complementing or even rivaling gains from scaling model size~\cite{Snell2024Scaling,Huang2025m1}. A dominant line of work samples multiple solutions and aggregates them via consensus (e.g., Self-Consistency) or selection (e.g., Best-of-M with a reward model)~\cite{wang2022self,Snell2024Scaling}. Beyond selection-only schemes, TTRL enables parameter adaptation on unlabeled inputs by deriving proxy rewards from repeated sampling and MV, yielding \textit{self-evolution} at test time~\cite{zuo2026ttrl,liu2025ettrl,pan2026coverrl}. While recent frontier reasoning models leverage outcome-verifiable rewards or expensive human annotations for RL~\cite{guo2025deepseek,jaech2024openai}, unlabeled test-time adaptation is particularly appealing in medical settings where real-time gold standards are rarely available~\cite{Hager2024Evaluation}. 

\vspace{0.6em}
\noindent \textbf{Inference-Time Verification: From Outcome to Process Rewards.} Test-Time Scaling improves output quality without updating model parameters, typically by exploring multiple reasoning trajectories and selecting the most reliable one. Outcome-based reward models score whole solutions but can misjudge trajectories due to spurious reasoning that happens to reach the correct answer or correct early steps that later derail~\cite{Yuan2025Curing,lee2025rethinking}. Process reward models address this by providing stepwise supervision to evaluate intermediate reasoning validity~\cite{Lightman2023Let's,Zheng2025A}. In medicine, stepwise verification is challenging due to the lack of symbolic verifiers; Med-PRM tackles this by grounding step evaluations in retrieved clinical guidelines and medical literature (RAG-as-a-judge), enabling stronger verifier-based selection in medical reasoning~\cite{Yun2025Med-PRM,Liu2025Improving}.





\section{Motivation: TTRL Fails on Medical MCQ}
\label{sec:diagnosis}

\vspace{2mm}
\noindent \textbf{Observed Collapse.} We first examine standard majority vote TTRL on medical multiple choice QA. As shown in Figure~\ref{fig:teaser}, TTRL produces only limited gains in sample-level accuracy, while majority vote accuracy remains essentially unchanged. At the same time, output diversity drops sharply~\cite{cui2025entropy,chen2026does}. The model therefore becomes more self-consistent without becoming more correct.
This behavior exposes a weakness of answer-level self-training. TTRL rewards agreement with the majority answer, but in medical MCQ, agreement is not a reliable proxy for correctness. When the answer space is small, different flawed reasoning paths can easily converge on the same option. Training then reinforces the shared answer even when the reasoning behind it is wrong.

\vspace{2mm}
\noindent \textbf{Answer Space Matters.} One possible explanation is that medical questions are simply harder than the mathematics tasks where TTRL has succeeded. To test this, we construct a controlled experiment on AMC mathematics problems~\cite{zuo2026ttrl}. Each question is evaluated in two formats: its original free-form format and a matched four-option multiple-choice format. The questions, policy model, optimizer, and training procedure are held fixed. Only the answer space changes.

\begin{figure}[t]
  \centering
  \includegraphics[width=\columnwidth]{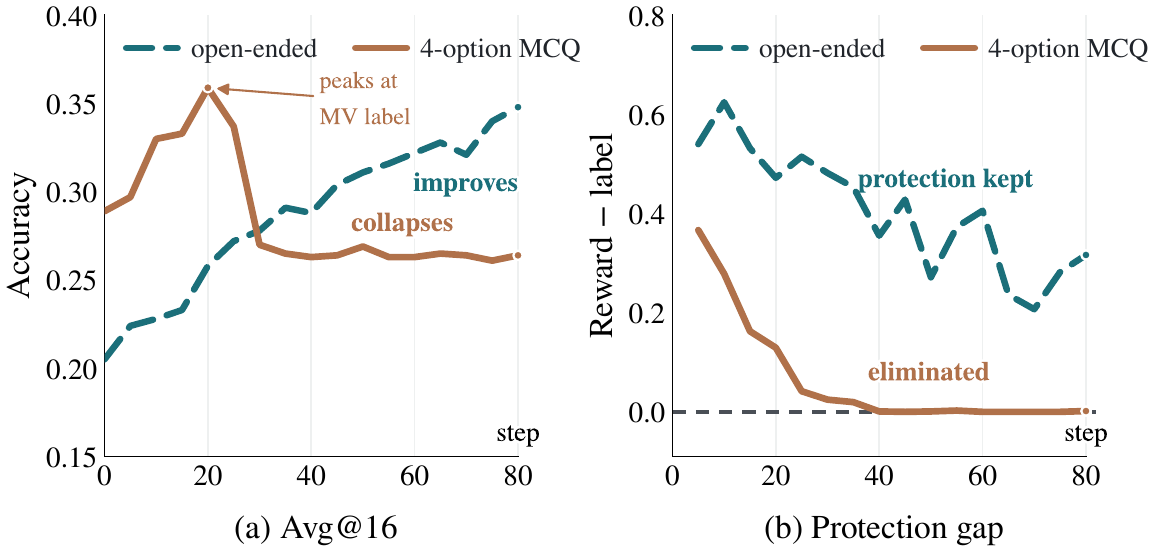}
\caption{Changing only the answer space flips TTRL from improvement
to collapse. \textbf{(a)} Avg@16 improves for open-ended answers but collapses
for 4-option MCQs. \textbf{(b)} The protection gap
(reward accuracy\ $-$ label accuracy) remains positive for open-ended answers but
vanishes for 4-option MCQs.}
  \label{fig:causal}
\end{figure}

Figure~\ref{fig:causal} shows that this change is sufficient to reverse the training dynamics. In the free form setting, TTRL steadily improves accuracy. In the matched multiple choice setting, accuracy initially rises but later collapses. Since the underlying questions and optimization procedure are unchanged, the result points to answer space structure rather than domain difficulty as a key cause of failure.

\vspace{2mm}
\noindent \textbf{The Lucky Hit Mechanism.} The controlled experiment reveals
how the \emph{Lucky Hit} effect depends on the answer space. In a large free-form answer space, incorrect rollouts are dispersed across many distinct strings. Even when the majority pseudo label is wrong, most other wrong rollouts do not exactly match it and therefore receive no reward. This sparsity protects training from many pseudo label errors.
In a small multiple-choice answer space, this protection disappears. Different mistaken rationales are compressed into the same few options, so wrong rollouts often collide on the same incorrect answer. Once that answer becomes the majority pseudo label, many flawed trajectories receive positive reward. The measured protection gap consequently falls from strongly positive in the free form setting to nearly zero under multiple choice.
This mechanism also explains why diversity collapse alone is not the root cause. Diversity can decrease while accuracy improves, as in the free form condition. What matters is where the probability mass concentrates. In medical MCQ, the small answer space allows probability mass to concentrate on a wrong option, turning a pseudo-label error into a self-reinforcing training signal.

\vspace{2mm}
\noindent \textbf{From Diagnosis to Reward Design.}  The diagnosis suggests that the reward should not depend only on final answer agreement. In collision-prone answer spaces, a more reliable signal should evaluate the reasoning process that produces the answer. This motivates PROSE, which replaces majority-vote rewards with process-level rewards from a medical PRM.
PROSE scores each reasoning step and assigns the trajectory reward as the minimum score across steps. This design makes the reward sensitive to the weakest step in the reasoning chain, so a response cannot hide a flawed inference behind fluent or easy steps. The PRM is used only during adaptation. Afterward, the adapted policy answers without PRM scoring, meaning the process signal is absorbed into the model rather than used as an inference-time selector.

\section{Method}
\label{sec:method}

\begin{figure*}[t]
  \centering
  \includegraphics[width=\textwidth]{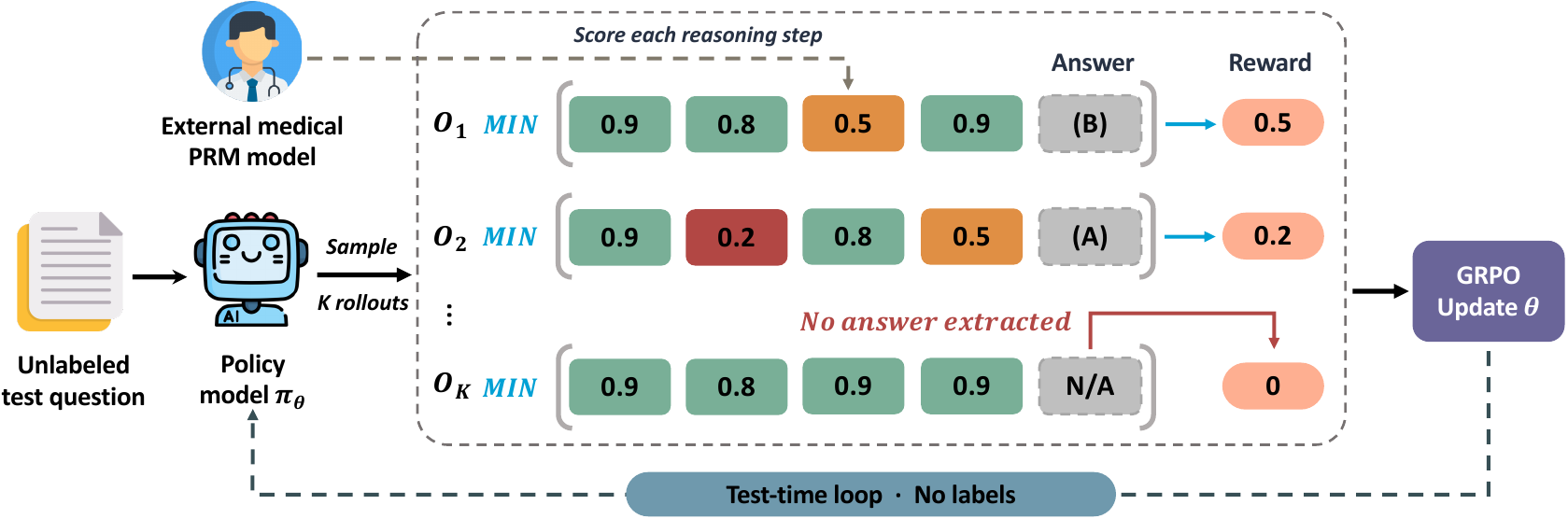}
\caption{\textbf{Overview of PROSE.} Given an unlabeled test question, the
policy $\pi_\theta$ samples $K$ rollouts. An external medical process reward
model scores each reasoning step, and the minimum step score defines the reward
for each rollout. A format guard assigns zero reward when no answer can be
extracted. GRPO then updates the policy using the rollouts and their rewards,
and the process repeats on the test set without labels.}
  \label{fig:method}
\end{figure*}

The preceding diagnosis shows that majority-vote TTRL fails because final-answer agreement becomes unreliable in a small answer space. PROSE (\textbf{Pro}cess-Reward-Guided \textbf{Se}lf-Training) keeps the same test-time RL loop but replaces answer-level pseudo-label rewards with process-level rewards. As shown in Figure~\ref{fig:method}, PROSE samples rollouts from the current policy, scores their reasoning steps with a medical process reward model (PRM), aggregates step scores with a minimum operator, applies a format guard, and updates the policy with GRPO. The entire procedure uses only unlabeled test questions.

\vspace{2mm}
\noindent \textbf{Test-Time RL Setup.} We consider transductive test-time adaptation on an unlabeled test set $\mathcal{D}=\{q\}$. For each question $q$, the policy samples a group of $K$
rollouts,
\begin{equation}
    \{o_k\}_{k=1}^{K}\sim\pi_\theta(\cdot\mid q),
\end{equation}
where each rollout $o_k=(s^{(k)}_1,\dots,s^{(k)}_{n_k},a_k)$ contains a sequence of reasoning steps followed by a final answer $a_k$.

Given rollout rewards $\{r_k\}_{k=1}^{K}$, we optimize the policy with GRPO~\cite{shao2024deepseekmath}. The advantage of rollout $k$ is normalized within the sampled group,
\begin{equation}
    A_k=\frac{r_k-\mu}{\sigma},
\end{equation}
where $\mu$ and $\sigma$ are the mean and standard deviation of the group rewards. GRPO then updates $\theta$ with the clipped policy-gradient objective.

Standard TTRL defines reward by matching the majority-vote answer,
\begin{equation}
r_{k,\mathrm{TTRL}} =
\begin{cases}
1, & \text{if } a_k=\hat{a},\\
0, & \text{otherwise},
\end{cases}
\qquad
\hat{a}=\operatorname{MV}\!\left(\{a_j\}_{j=1}^{K}\right).
\end{equation}
This reward depends only on agreement in answer space, while PROSE replaces this signal with a reward over reasoning steps.

\vspace{2mm}
\noindent \textbf{Process-Level Reward.} PROSE evaluates each rollout by the quality of its reasoning process rather than by agreement with other sampled answers. A medical PRM assigns a correctness score to each reasoning step,
\begin{equation}
    p^{(k)}_i =
    \mathrm{PRM}\!\left(s^{(k)}_i \mid q, s^{(k)}_{<i}\right)\in[0,1].
\end{equation}
This score is independent of whether other rollouts choose the same final option, so it is not directly affected by answer-space collisions.

To convert step scores into a trajectory reward, PROSE uses the minimum score across reasoning steps,
\begin{equation}
    r_k=\min_i p^{(k)}_i.
\end{equation}
This aggregation matches the structure of multi-step reasoning: a chain is only as reliable as its weakest step. It also limits reward hacking~\cite{pmlr-v202-gao23h,Skalse2022DefiningAC,Amodei2016ConcretePI}. Mean or sum aggregation can be increased by adding high-scoring steps that dilute a flawed one, whereas the minimum can improve only when the weakest step improves. Empirically, this distinction is important: mean aggregation saturates the proxy
reward while degrading accuracy, whereas minimum aggregation remains aligned with answer correctness.

\vspace{2mm}
\noindent \textbf{Format Guard.} The PRM scores reasoning steps but does not evaluate whether the rollout ends with a valid final answer. Without an explicit constraint, optimization can favor high-scoring reasoning traces that fail to produce a parseable answer. PROSE therefore applies a format guard:
\begin{equation}
  r_k =
  \begin{cases}
    \min_i p^{(k)}_i, & a_k \text{ is parseable},\\[2pt]
    0, & \text{otherwise}.
  \end{cases}
\end{equation}
A rollout receives high reward only when it contains both a valid final answer and consistently reliable reasoning steps.

\vspace{2mm}
\noindent \textbf{Algorithm and Inference Cost.} Algorithm~\ref{alg:main} summarizes one PROSE update. The PRM is used only during test-time adaptation to provide process-level rewards. After adaptation, the policy is evaluated directly with ordinary sampling or majority voting and requires no PRM scoring at inference. Thus, PROSE uses the PRM as a temporary training signal rather than as a per-query selector.
\begin{algorithm}[t]
\caption{PROSE, one optimization step}
\label{alg:main}
\small
\begin{tabular}{@{}r@{~~}p{0.82\columnwidth}@{}}
\multicolumn{2}{@{}l}{\textbf{Input:} question $q$, policy $\pi_\theta$, process model \textsc{Med-PRM}}\\[2pt]
1 & Sample $K$ rollouts $\{o_k\}_{k=1}^{K}\sim\pi_\theta(\cdot\mid q)$ \\
2 & \textbf{for} each rollout $o_k$ \textbf{do} \\
3 & \quad Score steps $p^{(k)}_i \leftarrow \textsc{Med-PRM}(s^{(k)}_i)$ \\
4 & \quad Set $r_k \leftarrow \min_i p^{(k)}_i$ \\
5 & \quad \textbf{if} $a_k$ is not parseable \textbf{then} $r_k\leftarrow 0$ \\
6 & \textbf{end for} \\
7 & Update $\theta$ with GRPO using rewards $\{r_k\}_{k=1}^{K}$ \\
\end{tabular}
\end{algorithm}

\section{Experiments Setup}
\label{sec:experiments setup}

\begin{table*}[t]
  \centering
\caption{Avg@16 performance on four medical QA datasets. Our method consistently
improves standard TTRL across all datasets. Best and second-best results among
models with fewer than 32B parameters are shown in \textbf{bold} and
\underline{underlined}, respectively.}
  \label{tab:matrix}
  \begin{tabular}{llcccccc}
    \toprule
    \textbf{Category} & \textbf{Model} & \textbf{Size} & \textbf{MedQA-5op} & \textbf{MedMCQA} & \textbf{DDXPlus} & \textbf{MedQA-4op} & \textbf{Avg.} \\
    \midrule
    \multirow{6}{*}{\makecell[l]{Large-scale \\General-purpose LLMs}}
    & Gemini Flash 2.0  & --       & 0.852 & 0.726 & 0.750 & 0.875 & 0.801 \\
    & GPT-4o-mini       & --       & 0.743 & 0.682 & 0.760 & 0.790 & 0.744 \\
    & GPT-3.5 turbo     & --       & 0.654 & 0.570 & 0.738 & 0.699 & 0.665 \\
    & QWQ               & 32B       & 0.769 & 0.609 & 0.761 & 0.825 & 0.741 \\
    & R1-Distill-Qwen   & 32B       & 0.738 & 0.591 & 0.781 & 0.772 & 0.721 \\
    & Sky-T1            & 32B       & 0.749 & 0.587 & 0.776 & 0.795 & 0.727 \\
    \midrule
    \multirow{4}{*}{\makecell[l]{Open-source Small \\ Language Models}}
    & Llama3.1          & 8B       & 0.636 & 0.556 & 0.695 & 0.654 & 0.635 \\
    & Gemma2            & 9B       & 0.532 & 0.512 & 0.676 & 0.581 & 0.575 \\
    & Ministral         & 8B       & 0.475 & 0.469 & 0.606 & 0.530 & 0.520 \\
    \midrule
    \multirow{7}{*}{\makecell[l]{Open-source Medical \\ Language Models}}
    & Med42             & 8B       & 0.575 & 0.573 & 0.597 & 0.614 & 0.590 \\
    & Meditron3         & 8B       & 0.569 & 0.549 & 0.632 & 0.612 & 0.591 \\
    & OpenBioLLM        & 8B       & 0.363 & 0.459 & 0.438 & 0.373 & 0.408 \\
    & m1-7B         & 7B       & \underline{0.681} & 0.568 & 0.662 & \underline{0.721} & 0.658 \\
    & Meerkat           & 8B       & 0.654 & 0.569 & 0.709 & 0.672 & 0.651 \\
    & UltraMedical      & 8B       & 0.665 & 0.579 & 0.694 & 0.710 & 0.662 \\
    & HuatuoGPT-o1      & 8B       & 0.666 & \underline{0.619} & 0.660 & 0.714 & 0.665 \\
    \midrule
    \multirow{2}{*}{\makecell[l]{Test-time Training \\on Llama3.1-8B}}
    & TTRL              & 8B       & 0.672 & 0.573 & \underline{0.761} & 0.676 & \underline{0.671} \\
    & \cellcolor{gray!25}Ours   & \cellcolor{gray!25}8B    & \cellcolor{gray!25}\textbf{0.724} 
                                                           & \cellcolor{gray!25}\textbf{0.623} 
                                                           & \cellcolor{gray!25}\textbf{0.858} 
                                                           & \cellcolor{gray!25}\textbf{0.756} 
                                                           & \cellcolor{gray!25}\textbf{0.740}  \\
    \bottomrule
  \end{tabular}
\end{table*}

\paragraph{Policies.} We evaluate two model families and three sizes:
Llama-3.1-8B-Instruct~\cite{grattafiori2024llama} and Qwen3-\{1.7B, 4B, 8B\} (with thinking disabled)~\cite{yang2025qwen3}. The
two 8B models are our main policies; the smaller Qwen3 sizes are used to study
the effect of model scale.
 
\paragraph{Process reward model.} The reward comes from Med-PRM~\cite{Yun2025Med-PRM}, an 8B medical
process reward model that assigns each reasoning step a probability of being
correct, $p_i(+)\in[0,1]$. Med-PRM is used only during test-time training: it provides the per-step scores that PROSE
aggregates into a reward; the adapted policy answers
at inference without it.
 
\paragraph{Datasets and protocol.} We use four clinical multiple-choice
benchmarks: MedQA-5op (5-option), MedQA-4op (4-option)~\cite{jin2021disease}, MedMCQA~\cite{pmlr-v174-pal22a}, and DDXPlus~\cite{fansi2022ddxplus}. Following the test-time RL setting, each run is transductive: the training set is the test set and no ground-truth labels are used at any point. A separate run
adapts each policy on each dataset.
 
\paragraph{Baselines.} We compare against: 
(i) the base model with no adaptation; 
(ii) TTRL, test-time RL with the majority-vote reward~\cite{zuo2026ttrl}; 
(iii) inference-time Med-PRM selection at a matched sampling budget,
including best-of-$N$ reranking (BoN) and reward-weighted
self-consistency (SC+RM);
and (iv) Large-scale General-purpose LLMs, open-source small language models and medical models re-run under our protocol~\cite{chen2024huatuogpt,zhang2024ultramedical,team2024gemini}.
 
\paragraph{Metrics.}
Our primary metric is avg@16: the average accuracy
over $16$ samples per question. Where we compare against inference-time selection (BoN, SC+RM), whose
output is a single chosen answer, we instead report maj@16, the accuracy
of the majority vote over $16$ samples, so that all methods spend the same
sampling budget. 
All models, including the external baselines, are evaluated identically: $16$ samples per question at temperature $0.6$ and top-$p$ $0.95$.
 
\paragraph{Optimization.} We optimize with GRPO, normalizing each
rollout's advantage by the group standard deviation. Each
update step uses $7$ prompts $\times\,32$ samples ($224$ rollouts), and we train
for $10$ epochs. We use AdamW (learning rate
$5\times10^{-7}$, cosine schedule with $3\%$ warmup, weight decay $0.01$,
$\beta=(0.9,0.999)$), a symmetric PPO clip of $0.2$ (no clip-higher) with one
PPO epoch and gradient clipping at $1.0$, a KL penalty (low-variance estimator,
coefficient $10^{-3}$), and no entropy bonus. Training rollouts use temperature
$1.0$ (top-$p$ $1.0$); prompts and responses are capped at $1024$ and $2048$
tokens (context $3072$). We do all experiments on A100 GPUs.

\section{Results and Analysis}
\label{sec:results}
\subsection{Main Results}
\noindent \textbf{PROSE lifts an 8B model to the top of its class.}
Table~\ref{tab:matrix} reports avg@16 across the four clinical
datasets, with every model evaluated under the same protocol. Applied to
Llama-3.1-8B, PROSE achieves an average score of $0.740$, the best result
among all 8--9B general and medical models. This represents a substantial
gain over its base model, from $0.635$ to $0.740$, or $10.5$ percentage
points. Using the same backbone, PROSE outperforms majority-vote TTRL by
$6.9$ percentage points, improving from $0.671$ to $0.740$. This result
shows that the improvement comes from the process reward rather than
test-time training alone. Notably, PROSE also surpasses every
purpose-built open-source medical model, the strongest of which achieves
approximately $0.665$, despite using a general model adapted without
labels.
 
Beyond its weight class, the 8B model adapted with PROSE is competitive
with much larger and proprietary systems. Its average score of $0.740$
nearly matches QwQ at $0.741$, the strongest 32B open reasoning model,
while using roughly one quarter of the parameters. It is also comparable
to GPT-4o-mini at $0.744$ and trails only Gemini Flash 2.0 at $0.801$.
Among models of comparable size, PROSE performs best on all four
datasets. On DDXPlus, it achieves $0.858$, the highest score in the entire
table, surpassing every proprietary and 32B model. Proprietary and 32B
models retain an advantage on the exam-style MedQA and MedMCQA
datasets, but PROSE closes much of this gap using a small, open model
and no labeled data.

\begin{table*}[t]
  \centering
\caption{Comparison with TTRL and PRM-based inference methods on four
medical QA datasets. PROSE achieves the best average performance for both
model families and the best result in seven of eight model-dataset settings.
Best results are shown in \textbf{bold}.}
  \label{tab:vs-inference}
  \footnotesize
  \setlength{\tabcolsep}{8pt}
  \begin{tabular}{llcccccc}
    \toprule
    \textbf{Model} & \textbf{Test-Time Method} & \textbf{MedQA-5op} & \textbf{MedMCQA} & \textbf{DDXPlus} & \textbf{MedQA-4op} & \textbf{Avg.}\\
    \midrule
    \multirow{5}{*}{Llama3.1-8B}
    & TTRL (maj@16)                   & 0.684 & 0.572 & 0.760 & 0.714 & 0.683 \\
    & PRM (BON@16)                    & 0.690 & 0.610 & 0.780 & 0.720 & 0.700 \\
    & PRM (SC+RM@16)                  & 0.690 & 0.620 & 0.740 & 0.740 & 0.698 \\
    & \cellcolor{gray!25}PROSE (maj@16)        & \cellcolor{gray!25}\textbf{0.740} 
                                              & \cellcolor{gray!25}\textbf{0.628} 
                                              & \cellcolor{gray!25}\textbf{0.863} 
                                              & \cellcolor{gray!25}\textbf{0.764}
                                              & \cellcolor{gray!25}\textbf{0.749}\\
    \midrule
    \multirow{4}{*}{Qwen3-8B}
    & TTRL (maj@16)                   & 0.696 & 0.606 & 0.866 & 0.729 & 0.724 \\
    & PRM (BON@16)                    & 0.690 & 0.590 & 0.810 & \textbf{0.770} & 0.715 \\
    & PRM (SC+RM@16)                  & 0.720 & 0.580 & 0.840 & 0.760 & 0.725 \\
    & \cellcolor{gray!25}PROSE (maj@16)     & \cellcolor{gray!25}\textbf{0.754} 
                                              & \cellcolor{gray!25}\textbf{0.645} 
                                              & \cellcolor{gray!25}\textbf{0.907} 
                                              & \cellcolor{gray!25}0.760 
                                              & \cellcolor{gray!25}\textbf{0.767}\\  
    \bottomrule
  \end{tabular}
\end{table*}

\noindent \textbf{Reshaping the policy beats reranking its samples.}
Med-PRM is designed for inference-time selection with BoN or SC+RM. We ask
whether the same reward model is more effective when used for selection or
test-time training. Since BoN and SC+RM return a single answer, we evaluate
PROSE using maj@16 from the same $N{=}16$ samples, ensuring a comparable
single-answer evaluation and inference-time sampling budget.
Table~\ref{tab:vs-inference} compares these approaches on two 8B policies.
The comparison with TTRL directly isolates the effect of the reward signal,
as the two methods otherwise follow the same test-time RL framework. PROSE
improves the average by $6.6$ percentage points on Llama-3.1-8B
($0.683\!\to\!0.749$) and by $4.3$ points on Qwen3-8B
($0.724\!\to\!0.767$), outperforming TTRL on every dataset. PROSE also
substantially outperforms inference-time PRM selection, achieving the best
result in $7$ of the $8$ dataset$\times$model settings. Compared with the
strongest PRM selection baseline, it improves the average from $0.700$ to
$0.749$ on Llama-3.1-8B and from $0.725$ to $0.767$ on Qwen3-8B. The largest
gains occur on DDXPlus, where PROSE reaches $0.863$ with Llama-3.1-8B and
$0.907$ with Qwen3-8B.
 
These results speak to how the PRM is best used. BoN and SC+RM can only pick among samples drawn from a fixed model, whereas PROSE uses the same Med-PRM to reshape the policy's output distribution, making correct reasoning more likely in the first place; at a matched sampling budget, reshaping beats reranking. The reward is thereby \emph{internalized} into the weights rather than merely consulted---the adapted policy answers by plain majority voting, with no PRM at inference, while BoN and SC+RM query the PRM on every question. We do not claim a net compute saving: test-time training is itself expensive. The point is that the PRM is needed only once, during adaptation, rather than on every future
query, so its cost amortizes whenever the adapted policy is reused.
We test this internalization next through out-of-distribution generalization.
 
\begin{figure}[t]
  \centering
  \includegraphics[width=\columnwidth]{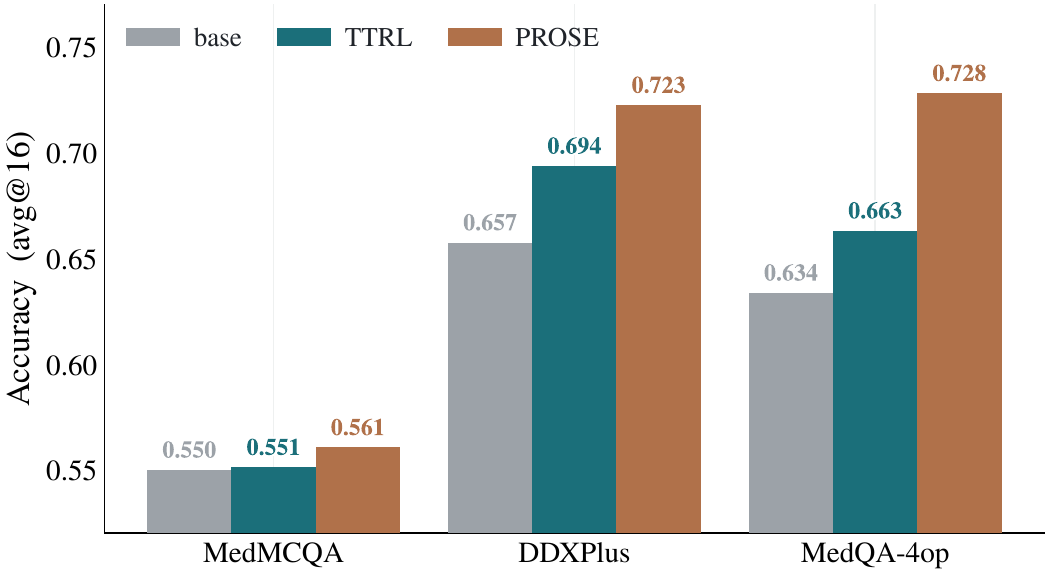}
\caption{Out-of-distribution generalization measured by avg@16.
The policy is adapted on MedQA-5op, then frozen and evaluated on three unseen
datasets without further adaptation or PRM scoring. PROSE consistently
outperforms both the base model and standard TTRL.}
  \label{fig:ood}
\end{figure}

\noindent \textbf{Out-of-distribution generalization.}
If PROSE merely fit the specific questions used for adaptation, its gains
would not transfer across datasets. We therefore freeze the policy adapted
on MedQA-5op and evaluate it on three datasets unseen during adaptation,
using standard sampling without the PRM. Figure~\ref{fig:ood} shows that
PROSE improves over the base model on all three out-of-distribution datasets
and outperforms TTRL in every case: MedMCQA improves from $0.550$ to $0.561$,
DDXPlus from $0.657$ to $0.723$, and MedQA-4op from $0.634$ to $0.728$.
As expected, these transferred gains are smaller than those obtained by
adapting directly to each target dataset (see Table~\ref{tab:matrix}).
The gain is largest on MedQA-4op, which most closely resembles the
adaptation set, and smallest on MedMCQA.
 
This is the clearest evidence that the process reward has been internalized. The
adapted policy carries its improvement to new distributions without any
PRM at inference.
The external evaluator was needed only once, to bootstrap the
policy, and what remains is a self-contained model rather than a set of
memorized answers.

\subsection{Ablation: What matters in the reward design}
\label{sec:exp-reward}
 
Two design choices decide how much of the process signal survives: how the
signal enters training, and how the per-step scores are aggregated.
 
\paragraph{A process reward beats a process label.}
Table~\ref{tab:reward-design} isolates the first choice on MedQA-5op.
Replacing majority voting with a PRM-derived pseudo-label already improves avg@16 from $0.672$ to $0.709$, confirming that the
process signal captures information missed by majority voting.
But keeping the PRM as a continuous
reward is better still ($0.724$ avg@16), and the gap widens under maj@16. The reason is what each
formulation discards: collapsing the step scores into a single target to imitate
throws away their gradation, so every incorrect trajectory looks equally wrong,
whereas a graded reward preserves the ordering among partially-correct
trajectories and tells the policy which of its attempts came closer.

\begin{table}[t]
  \centering
\caption{Effect of incorporating process supervision into TTRL on MedQA-5op with
Llama3.1-8B. PRM-derived labels outperform majority-vote labels, while using
PRM scores as graded rewards (PROSE) achieves the best avg@16 and maj@16
performance. Best results are shown in \textbf{bold}.}
  \label{tab:reward-design}
  \small
  \setlength{\tabcolsep}{8pt}
  \begin{tabular}{@{}lcc@{}}
    \toprule
    Method & avg@16 & maj@16 \\
    \midrule
    MV as label  & 0.672 & 0.684 \\
    PRM as label & 0.709 & 0.710 \\
    \textbf{PRM as reward (PROSE)} & \textbf{0.724} & \textbf{0.740} \\
    \bottomrule
  \end{tabular}
\end{table}

\begin{table*}[t]
  \centering
\caption{Effect of model scale on maj@16 performance. PROSE performs best
across all four datasets at 4B, while at 1.7B its advantage is limited to
DDXPlus, indicating that effective process-reward training depends on
sufficient policy capacity. Best results within each model block are shown
in \textbf{bold}.}
  \label{tab:scale}
  \footnotesize
  \setlength{\tabcolsep}{8pt}
  \begin{tabular}{llcccccc}
    \toprule
    \textbf{Model} & \textbf{Test-Time Method} & \textbf{MedQA-5op} & \textbf{MedMCQA} & \textbf{DDXPlus} & \textbf{MedQA-4op} & \textbf{Avg.} \\ 
    \midrule
    \multirow{4}{*}{Qwen3-4B}
    & TTRL (maj@16)           & 0.626 & 0.512 & 0.814 & 0.657 & 0.652 \\ 
    & PRM (BON@16)            & 0.680 & 0.570 & 0.790 & 0.700 & 0.685 \\
    & PRM (SC+RM@16)          & 0.670 & 0.570 & 0.800 & 0.690 & 0.683 \\
    & \cellcolor{gray!25}PROSE (maj@16)     & \cellcolor{gray!25}\textbf{0.725} 
                                              & \cellcolor{gray!25}\textbf{0.621} 
                                              & \cellcolor{gray!25}\textbf{0.856} 
                                              & \cellcolor{gray!25}\textbf{0.733}
                                              & \cellcolor{gray!25}\textbf{0.734}\\  
    \midrule
    \multirow{4}{*}{Qwen3-1.7B}
    & TTRL (maj@16)           & 0.436 & 0.462 & 0.826 & 0.520 & 0.561 \\                          
    & PRM (BON@16)            & \textbf{0.570} & \textbf{0.570} & 0.800 & \textbf{0.590} & \textbf{0.633} \\
    & PRM (SC+RM@16)          & 0.530 & 0.510 & 0.810 & 0.570 & 0.605 \\
    & \cellcolor{gray!25}PROSE (maj@16)      & \cellcolor{gray!25}0.493
                                              & \cellcolor{gray!25}0.553 
                                              & \cellcolor{gray!25}\textbf{0.871} 
                                              & \cellcolor{gray!25}0.540
                                              & \cellcolor{gray!25}0.614\\  
    \bottomrule
  \end{tabular}
\end{table*}

\begin{figure}[t]
  \centering
  \includegraphics[width=\columnwidth]{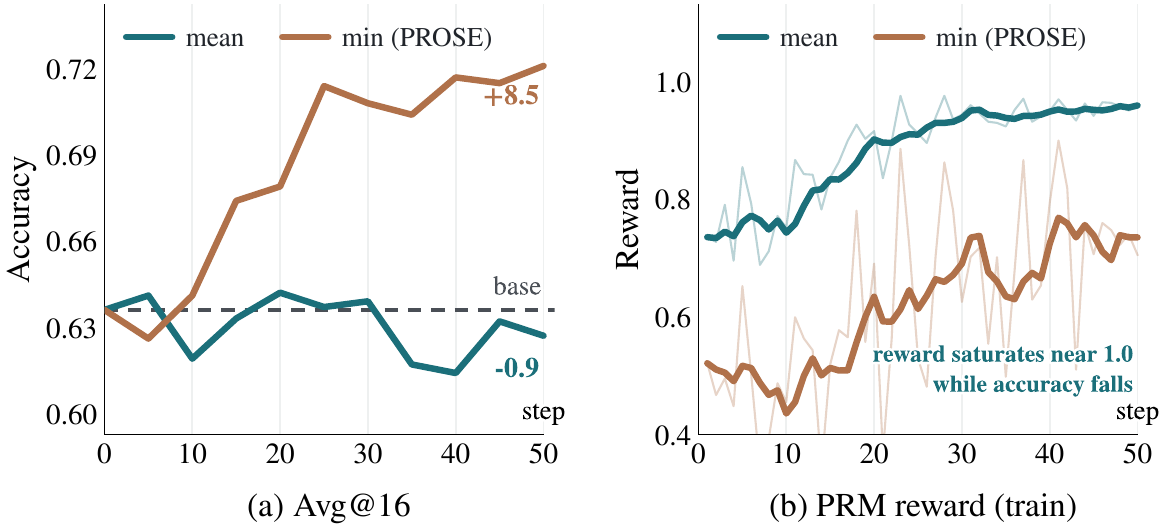}
\caption{\textsc{Min} vs.\ mean aggregation of step-level PRM scores on
MedQA-5op. \textbf{(a)} With \textsc{min} aggregation (PROSE), accuracy improves
and remains above the untrained base, whereas mean aggregation eventually
falls below it. \textbf{(b)} Despite the accuracy degradation, the
mean-aggregated PRM reward saturates near $1.0$, indicating reward hacking.
Faint lines show the raw per-step rewards, and thick lines show their
5-step moving averages.}
  \label{fig:minmean}
\end{figure}

\paragraph{\textsc{Min} aggregation keeps the reward honest.}
The second design choice is how to aggregate the step-level PRM scores,
which directly determines the method's susceptibility to reward hacking.
Figure~\ref{fig:minmean} compares minimum and mean aggregation under
otherwise identical settings. With minimum aggregation, avg@16 improves
by $8.5$ percentage points over the untrained base and remains consistently
above it. With mean aggregation, avg@16 eventually falls $0.9$ points below
the base, even as the optimized PRM reward approaches $1.0$. The proxy is
therefore nearly saturated while the true objective deteriorates, providing
clear evidence of reward hacking.
This behavior is consistent with the failure mode described in
Section~\ref{sec:method}. Mean aggregation allows low-scoring
reasoning steps to be diluted by additional high-scoring steps, enabling
the policy to increase its reward without correcting its weakest reasoning.
Minimum aggregation blocks this strategy because the trajectory reward
cannot exceed its lowest step-level score. Optimization must therefore
improve the weakest part of the trajectory. Accordingly, the minimum-
aggregated reward remains well below saturation, at approximately
$0.73$--$0.75$, while accuracy improves.
 
\subsection{Ablation: Effect of model scale}
\label{sec:exp-matrix}

Does internalizing the reward still pay off when the policy is small?
Table~\ref{tab:scale} compares PROSE with TTRL and inference-time PRM
selection on Qwen3-4B and Qwen3-1.7B under the same matched budget of
$N{=}16$ samples per question.

At 4B the picture is unchanged from 8B: PROSE is the best method on all four
datasets and by a wide margin on average ($0.734$ vs.\ $0.685$ for the strongest
selector), so the benefit of training with the process reward is not an artifact
of scale. At 1.7B, however, the ordering reverses on three of four
datasets---inference-time best-of-$N$ leads on average ($0.633$ vs.\
$0.614$)---and PROSE prevails only on DDXPlus.
 
This reversal is informative rather than incidental. Training with the process
reward requires the policy to be able to produce better reasoning once it
is pushed toward it; selection only requires that a good trajectory occasionally
appear among $N$ samples and be recognizable to the PRM. When the policy is very
weak, the second condition is easier to satisfy than the first, so the external
selector retains an advantage. 
The practical implication is that internalization
pays off once the policy is strong enough to act on the signal, from roughly 4B
in our setting, while below that scale the reward is better used at inference
time.

\section{Conclusion}
\label{sec:conclusion}

We showed that test-time RL, effective on mathematics, collapses on medical
multiple-choice QA for a structural reason: in a small answer space, incorrect
rollouts collide on the same wrong pseudo-label and reinforce it. A controlled
experiment varying only the answer space isolates this effect, and the
reward$-$label protection gap quantifies it ($+0.246$ free-form vs.\
$+0.001$ multiple-choice). PROSE follows from this diagnosis, rewarding the
reasoning process rather than answer agreement through minimum step-level
aggregation and an answer-format guard. Without labels, it gives
Llama-3.1-8B the best average accuracy among the evaluated $8$--$9$B models and
matches much larger systems. The adapted policy requires no reward model at
inference, outperforms PRM selection at the same sampling budget, and retains
its gains on unseen adaptation datasets.

\paragraph{Limitations and future work.} PROSE requires a domain-relevant
process reward model. Self-generated or general-purpose process rewards could
reduce this dependence and extend it beyond medical QA. Its effectiveness also
depends on policy capacity: inference-time selection remains preferable below
roughly 4B parameters in our experiments. Future work should characterize this
transition and how policy capacity and reward quality affect reward
internalization.

\bibliography{aaai2027}


\end{document}